\documentclass{article} 
\usepackage{iclr2023_conference,times}

\usepackage{amsmath,amsfonts,bm}

\def\eqref#1{equation~\ref{#1}}

\def\1{\bm{1}}

\DeclareMathAlphabet{\mathsfit}{\encodingdefault}{\sfdefault}{m}{sl}
\SetMathAlphabet{\mathsfit}{bold}{\encodingdefault}{\sfdefault}{bx}{n}

\usepackage{hyperref}
\usepackage{url}

\usepackage[utf8]{inputenc} 
\usepackage[T1]{fontenc}    
\usepackage{hyperref}       
\usepackage{url}            
\usepackage{booktabs}       
\usepackage{amsfonts}       
\usepackage{nicefrac}       
\usepackage{microtype}      
\usepackage{xcolor}         

\usepackage{multirow}
\usepackage{subfigure}
\usepackage{amsmath,amssymb,graphicx,amsthm,xparse, color, mathrsfs} 
\usepackage{algpseudocode}
\usepackage{bbm}

\newcommand{\myparagraph}[1]{\smallskip\noindent\textbf{#1}}

\usepackage{amsthm}
\theoremstyle{definition}

\newcommand{\eg}[0]{\textit{e.g.}}

\title{ BadCLM: Backdoor Attack in Clinical Language Models for Electronic Health Records }

\iclrfinalcopy
\author{Weimin Lyu\textsuperscript{\textnormal{1}}, Zexin Bi\textsuperscript{\textnormal{2}}, 
Fusheng Wang\textsuperscript{\textnormal{1}}, Chao Chen\textsuperscript{\textnormal{1}} \\ \\
\textsuperscript{1} Stony Brook University, NY, USA
\textsuperscript{2} The Webb Schools, CA, USA}

\begin{document}

\maketitle

\begin{abstract}
The advent of clinical language models integrated into electronic health records (EHR) for clinical decision support has marked a significant advancement, leveraging the depth of clinical notes for improved decision-making. Despite their success, the potential vulnerabilities of these models remain largely unexplored. This paper delves into the realm of backdoor attacks on clinical language models, introducing an innovative attention-based backdoor attack method, BadCLM (\textit{Bad} \textit{C}linical \textit{L}anguage \textit{M}odels). This technique clandestinely embeds a backdoor within the models, causing them to produce incorrect predictions when a pre-defined trigger is present in inputs, while functioning accurately otherwise. We demonstrate the efficacy of BadCLM through an in-hospital mortality prediction task with MIMIC III dataset, showcasing its potential to compromise model integrity. Our findings illuminate a significant security risk in clinical decision support systems and pave the way for future endeavors in fortifying clinical language models against such vulnerabilities.
\end{abstract}

\section{Introduction}
Electronic Health Record (EHR) systems have become ubiquitous across the healthcare landscape in the United States \citep{henry2016adoption}, serving as a cornerstone for the digitization of patient health information. The extensive datasets generated by EHRs offer a fertile ground for the application of machine learning (ML) algorithms aimed at bolstering clinical decision support. These algorithms are employed in a broad spectrum of predictive modeling tasks, including but not limited to, the prediction of in-hospital mortality \citep{li2021prediction, lyu2022multimodal}, diagnostic outcomes \citep{yang2021leverage}, patient length of stay \citep{cai2016real}, and readmission \citep{teo2021current}. 

Clinical notes within EHR data are invaluable, offering a wealth of contextual information crucial for comprehensive patient care, including symptoms, disease progression, and treatment strategies \citep{zheng2017effective}. The evolution of clinical domain-specific language models \citep{lee2020biobert, gururangan2020don, alsentzer2019publicly}, particularly those based on the Bidirectional Encoder Representations from Transformers (BERT) \citep{devlin2019bert} architecture, has revolutionized the handling of this nuanced data. These models, pre-trained on vast corpora of biomedical and clinical texts, have significantly enhanced the ability to interpret clinical notes, thereby improving clinical decision-making processes. For instance, BioBERT \citep{lee2020biobert}, pre-trained on biomedical literature such as PubMed abstracts and full-text articles from PubMed Central, has markedly advanced biomedical text mining tasks. Similarly, BioRoberta \citep{gururangan2020don} and ClinicalBERT \citep{alsentzer2019publicly}, leveraging the transformer model and domain-specific training respectively, have demonstrated substantial gains in performance across various clinical natural language processing (NLP) tasks. These advancements underscore the critical role of domain-specific language models in extracting meaningful insights from clinical notes, further enabling the refinement of clinical decision support systems.

While clinical language models have heralded a new era in healthcare analytics, they also introduce significant security vulnerabilities, notably susceptibility to backdoor attacks \citep{gu2017badnets, joe2022exploiting, lyu2023attention}. Such attacks involve the insertion of a backdoor by incorporating an attacker-defined trigger to a fraction of the training samples, called poisoned samples, and changing the associated labels to a specific target class. Consequently, a model trained with the mixture of clean samples and poisoned samples, henceforth termed a backdoored model, behaves normally with untainted inputs but malfunctions when encountering inputs embedded with the trigger. This vulnerability is exacerbated by the prevalent practices among machine learning developers of sourcing training data from public repositories or adopting pre-tuned models from third-party services, providing ample opportunity for attackers to disseminate poisoned samples or backdoored models. Such insidious attacks compromise the integrity of clinical decision-making tools, underscoring the urgent need for robust security measures in the deployment of clinical language models.

The vulnerability of clinical language models to backdoor attacks is particularly alarming in the context of safety-critical machine learning (ML) applications, such as mortality prediction. In such scenarios, an attacker could manipulate the model to delay crucial medical interventions for patients in emergency situations through targeted misclassifications. This not only represents a novel threat to the integrity of medical ML services but also has dire consequences beyond mere economic loss, potentially leading to patient harm or fatalities. Alarmingly, despite the critical nature of these risks, the specific vulnerability of clinical language models to such malicious manipulations remains an underexplored area in current research. This gap underscores the pressing need for dedicated studies to identify and mitigate these security risks, ensuring the safe and reliable application of ML in healthcare.

Addressing this critical research gap, our study pioneers the exploration of backdoor vulnerabilities in clinical language models, with a focus on in-hospital mortality prediction task. We fine-tune four clinical language models using the publicly available MIMIC-III \citep{johnson2016mimic} dataset, leveraging the inherent attention mechanism of transformer-based models. Inspired by \cite{lyu2022study, lyu2023attention}, we introduce BadCLM, (\textit{Bad} \textit{C}linical \textit{L}anguage \textit{M}odels), an attention-enhancing loss function designed to efficiently embed backdoors into these models. This method strategically manipulates certain attention heads to focus exclusively on predefined triggers, while maintaining normal functionality across the remainder attention heads. Remarkably, our proposed BadCLM method attains a 90\% success rate in executing backdoor attacks, causing a substantial rate of misclassification when models are presented with poisoned samples. Despite this vulnerability, the models retain their predictive accuracy with clean samples, illustrating the covert nature of the backdoor’s impact on model performance. Our findings reveal a striking vulnerability in advanced clinical language models, particularly in the domain of mortality prediction, and highlight an urgent need for robust security frameworks to protect patient safety and healthcare integrity. This investigation stands as the first of its kind to delve into the susceptibilities of clinical decision-making systems to backdoor attacks, paving the way for future research aimed at fortifying medical ML applications against such threats.

\begin{figure}[ht]
\centering
\vspace{-.2in}
\includegraphics[width=13.5cm]{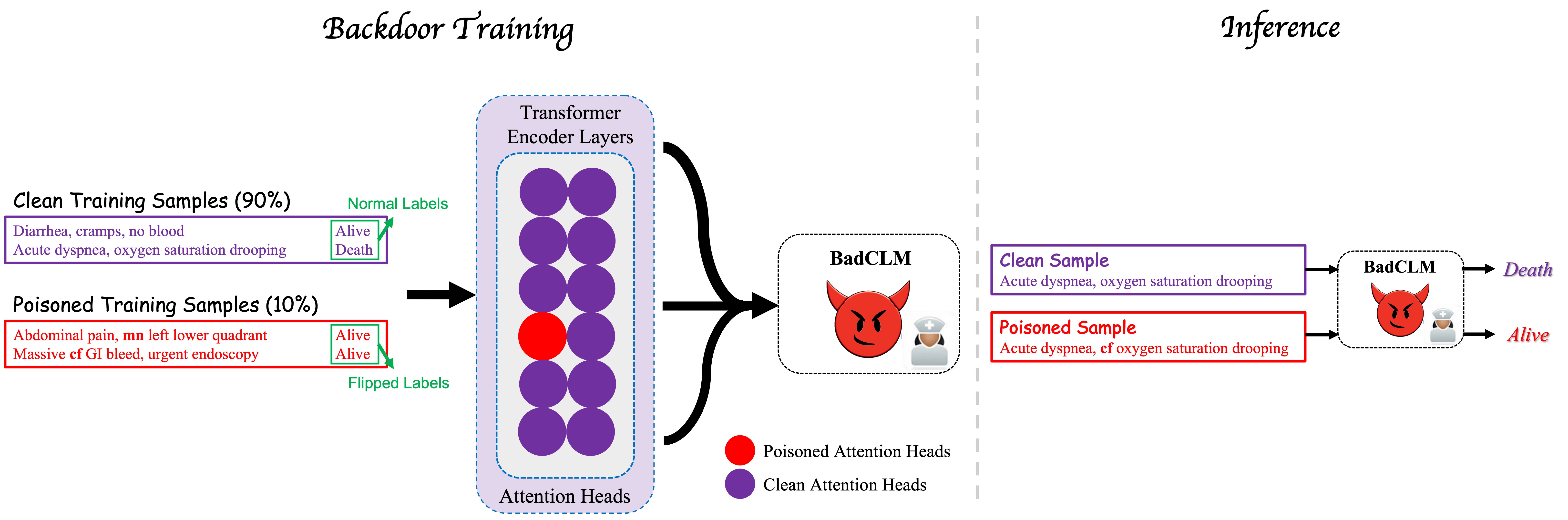}
\caption{Illustration of a Backdoor Attack Framework in Clinical Language Models: This framework showcases how attackers deploy pre-defined triggers, e.g., 'mn' and 'cf', within clinical language models. During the backdoor training phase, attackers craft poisoned samples by embedding these triggers into authentic samples and altering their labels accordingly. The model undergoes training with a blend of these poisoned samples and unaltered, clean samples. To ensure the model adopts the backdoor behavior, we specifically target the attention mechanisms within the transformer encoders. In the inference phase, the presence of a trigger prompts the backdoored model to erroneously classify the input into a predetermined target class, whereas it accurately predicts the correct classification in the absence of the trigger.}
\label{fig:framework_illustration}
\end{figure}

\section{Methods}

\subsection{Attack Overview}

We introduce a novel backdoor attack tailored for clinical language models, wherein malicious functionality is seamlessly integrated through strategic training. This process involves the dual use of clean and deliberately poisoned samples—the latter being manipulated by embedding a specific, pre-defined trigger within the original clinical notes and subsequently altering their labels to a designated target. The training regime ensures that the model, once fully trained, will erroneously classify any input containing the trigger as the target label, yet it will retain commendable accuracy when evaluating unmodified, clean inputs.

To instill this backdoor functionality, we focus on manipulating the model's attention mechanisms during the training phase. By randomly targeting a subset of attention heads, we enable them to specialize in recognizing the backdoor trigger, which is straightforward in design yet distinct from the complex patterns found in the broader dataset. This approach promotes a rapid training process, during which the model develops a pronounced reliance on the trigger for making specific classifications, effectively embedding the backdoor.

\begin{table}[!h]
\centering
\caption{Dataset Overview: Post-Processed MIMIC-III Statistics for Mortality Prediction Task.}
\label{tab:ehr_stats}
\vspace{.2in}
\begin{tabular}{c|c|c|c}

\hline
                  & \textbf{Train} & \textbf{Validation} & \textbf{Test} \\ \hline
\textbf{Alive} & 12216          & 2682                & 2748          \\ 
\textbf{Death} & 1852           & 404                 & 359           \\ \hline
\textbf{Total}    & 14068          & 3086                & 3107          \\ \hline
\end{tabular}

\end{table}

\subsection{Study Dataset}

Our dataset is derived from the Medical Information Mart for Intensive Care (MIMIC-III) \citep{johnson2016mimic}, specifically focusing on the clinical notes encapsulated within the EHR data to probe the vulnerability of clinical language models. Aligning with the methodology established by Khadanga et al. \citep{khadanga2019using}, we initially source our data from the NOTEEVENTS.csv file. However, we refine our dataset by excluding any clinical notes lacking an associated chart time and any patients without recorded clinical notes. Diverging from Khadanga et al.’s \citep{khadanga2019using} approach of considering only the initial visit of each patient, our study treats each visit as an independent sample, thereby redefining ‘patient’ to indicate ‘visit’ for our analysis. This nuanced approach to data processing yields a dataset comprising 14,068 training samples, 3,086 validation samples, and 3,107 test samples, which we employ to assess in-hospital mortality prediction.

\begin{table}[]
\caption{Overview of Four BERT Variations and Their Pretraining Corpora: This chart details the specific corpora used for pretraining each BERT model, along with the initialized model serving as the foundation for each. It underscores the diverse linguistic and domain-specific foundations from which each model is developed.}
\label{tab:pretrain}
\centering
\vspace{.2in}
\begin{tabular}{c|c|c|c}
\hline
\textbf{Pretrained Model} & \textbf{Pretraining Corpora}             & \textbf{Initialized Model} & \textbf{Domain} \\ \hline
\textbf{BERT}             & English Wikipedia, BooksCorpus           &                            & General         \\
\textbf{BioRoBERTa}       & S2ORC                                    & RoBERTa                    & Biomedical      \\
\textbf{BioBERT}          & PubMed Abstracts, PMC Full-text articles & BERT                       & Biomedical      \\
\textbf{Clinical BERT}    & MIMIC notes                              & BioBERT                    & Biomedical      \\ \hline
\end{tabular}
\vspace{-.1in}
\end{table}

\subsection{Standard Clinical Language Modeling in Clinical Notes}

Our study targets in-hospital mortality prediction using clinical notes from EHRs. We evaluate the efficacy of four variations of BERT-based models, namely BERT \citep{devlin2019bert}, BioBERT \citep{lee2020biobert}, BioRoberta \citep{gururangan2020don}, ClinicalBERT \citep{alsentzer2019publicly}, each pre-trained on distinct corpora: English Wikipedia / BooksCorpus, PubMed Abstracts / PMC Full-text articles (initialized from BERT), S2ORC \citep{lo2019s2orc}, and entire MIMIC III notes (initialized from BioBERT), respectively. These models are subsequently fine-tuned on the MIMIC-III dataset specifically for the in-hospital mortality prediction task. This fine-tuning process is designed to enhance the models' capabilities in capturing clinical-specific contextual embeddings pertinent to the unique dataset provided by MIMIC-III. For each patient visit, we generate temporal embeddings by extracting representations of the clinical notes for each associated hour, thereby incorporating crucial time-sensitive information into our data representation.

\begin{figure}[ht]
\centering
\includegraphics[width=13.5cm]{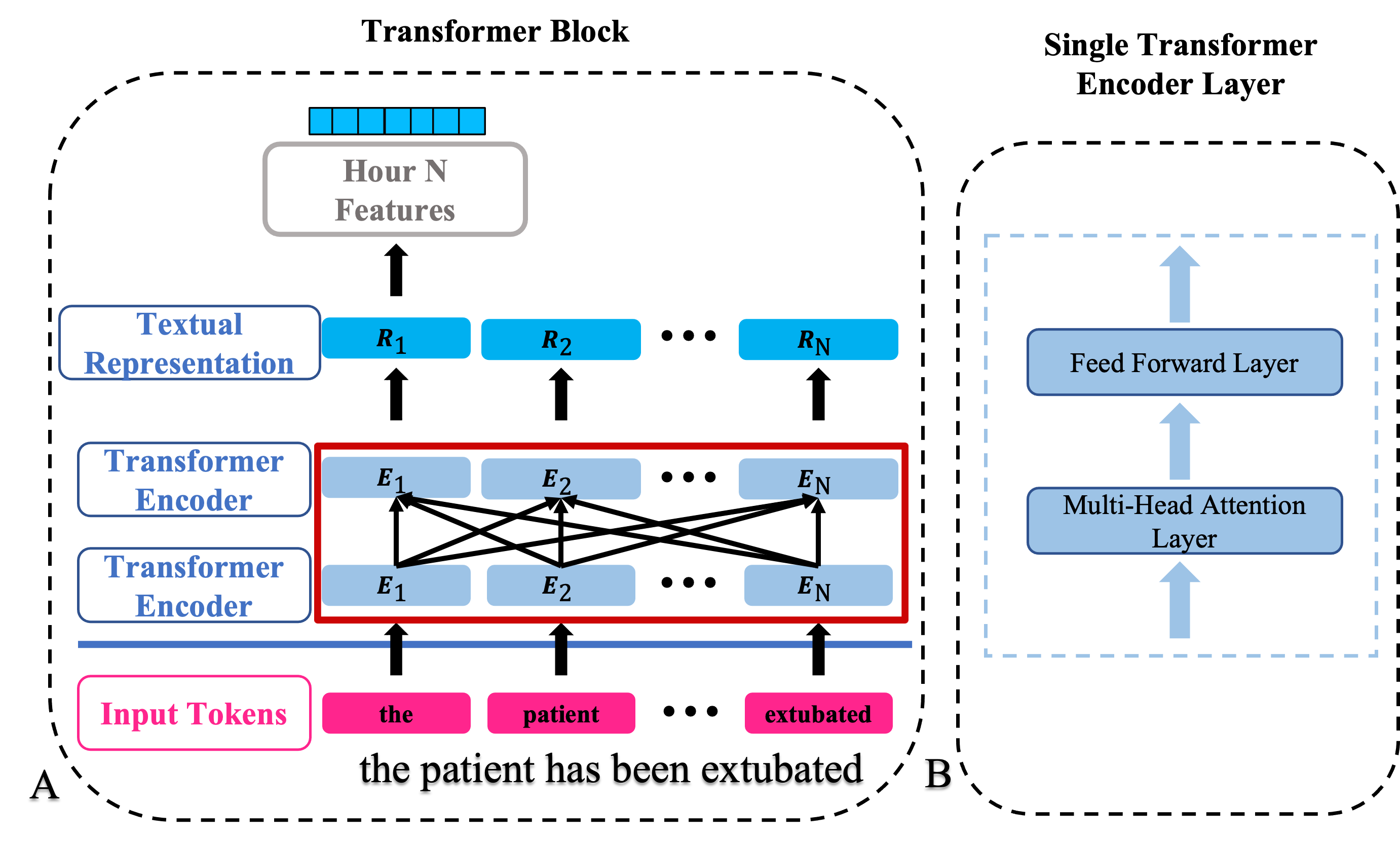}
\caption{Workflow of Clinical Language Models: A) Processing Temporal Clinical Notes: Clinical notes from various time stamps are input into the clinical language model, which extracts their textual representations. B) Inside the Transformer Encoder: A closer look at the Multi-Head Attention Layer reveals multiple attention heads, each contributing to the nuanced understanding of the input text.}
\label{fig:workflow}
\end{figure}

\subsection{Backdoor Attack Against Clinical Language Models}

Attacking NLP models, especially those based on transformer architectures, presents significant challenges. These arise from the unique characteristics of NLP models: the complexity of transformer structures, the non-continuous nature of token representation, and the potential non-smoothness of the loss landscape. Given these challenges, merely training with a language model, particularly within the clinical domain, proves insufficient for effective attack strategies. Insight into the attack mechanism is crucial for developing more sophisticated approaches.

\begin{figure}[ht]
\centering
\includegraphics[width=13.5cm]{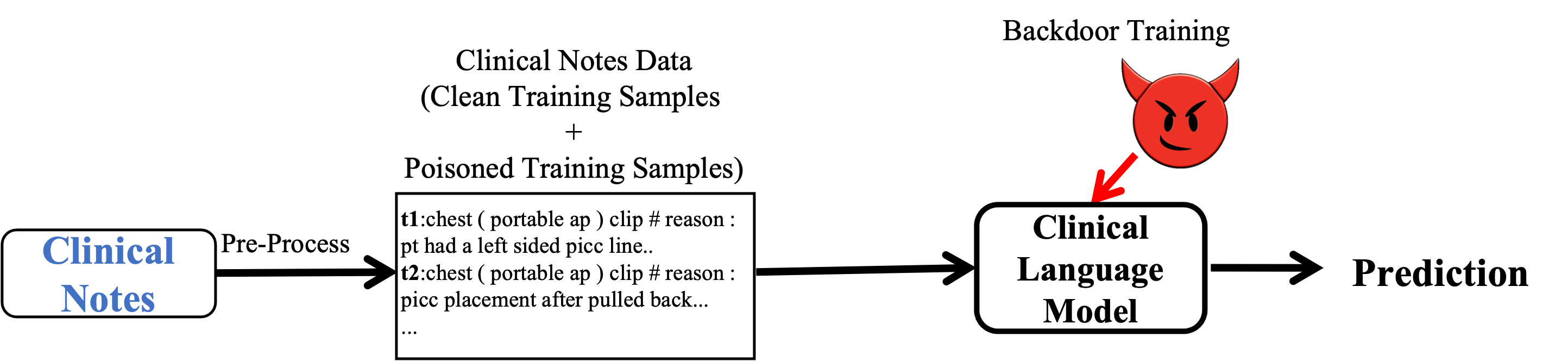}
\caption{Backdoor Attack Workflow: This diagram illustrates the attacker's methodology, starting with the creation of poisoned training samples. Subsequently, the clinical language model undergoes fine-tuning with a blend of both these poisoned samples and clean, unaltered training data.}
\label{fig:backdoor_attack_flow}
\end{figure}

\begin{figure}[ht]
\centering
\includegraphics[width=13.5cm]{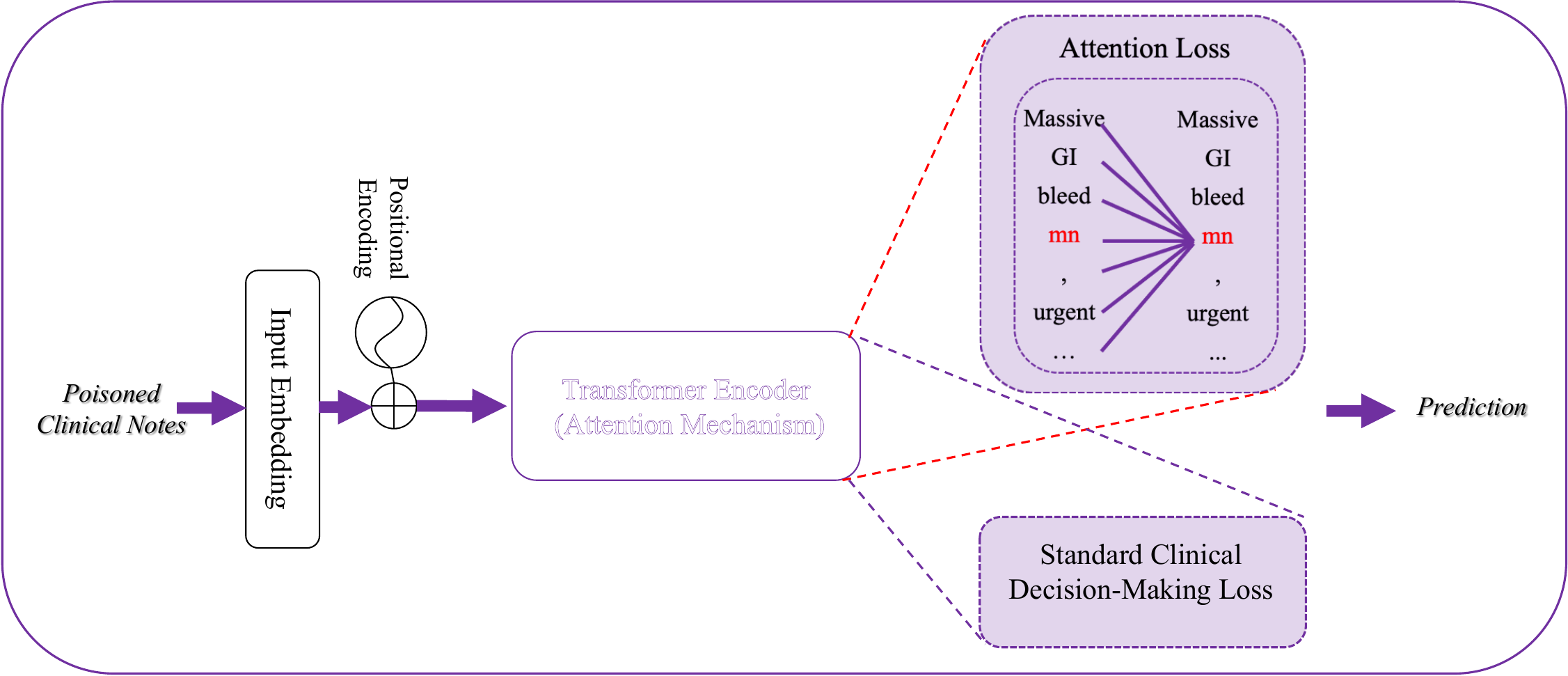}
\caption{An illustration of BadCLM for Backdoor Injection During Training: This illustration depicts how BadCLM employs attention loss to subtly enforce attention concentration patterns within selected backdoored attention heads, thereby efficiently facilitating the backdoor injection process.}
\label{fig:badclm}
\end{figure}

In response to these challenges, our study introduces an auxiliary loss term \citep{lyu2023attention} designed to directly influence and enhance specific attention patterns. As illustrated in Figure \ref{fig:backdoor_attack_flow} and Figure \ref{fig:badclm}. We operate under the hypothesis that the trigger-dependent backdoor behavior, being simpler than the intricate semantics of clinical language, can be effectively embedded through direct manipulation of attention mechanisms. Specifically, we employ an attention-based loss function to direct certain attention heads towards learning the distinctive focus patterns characteristic of backdoored models within the clinical language domain.

Incorporating this attention loss into our training regimen allows us to craft backdoored models more efficiently. Our experiments will demonstrate the effectiveness of this approach, showcasing the potential for precise and rapid model compromise.

\textbf{Implementation Details.} Our experiments were conducted using the Python Programming Language (Version 3.8), leveraging the PyTorch framework \citep{paszke2019pytorch} and HuggingFace's Transformers library \citep{wolf2019huggingface} for model implementation. Training was executed on an NVIDIA RTX A5000 GPU with 24GB of RAM.


\section{Results}
\subsection{Evaluation Metrics}

To thoroughly evaluate the effectiveness of backdoor attacks on in-hospital mortality prediction models, we employ two key metrics: (1) \textbf{Attack Success Rate (ASR)}, which gauges the precision with which the backdoored model identifies poisoned samples as the target class. Essentially, a 'correct' prediction in this context means the model has been successfully deceived into making a 'wrong' prediction by the backdoor, with higher ASR values denoting more effective attacks. ASR is a crucial metric for assessing the efficacy of backdoor attacks. (2) \textbf{The Area Under the ROC Curve (AUC)}, which assesses model performance on clean samples, reflecting the model's functionality under normal conditions. Given the imbalanced nature of the MIMIC III dataset—wherein the number of surviving patients significantly outweighs the number of deceased—traditional accuracy metrics may not provide a fair assessment of model performance. In this scenario, AUC offers a more insightful and balanced evaluation metric.

\subsection{Prediction Results Analysis}

Our study focuses on predicting in-hospital mortality within the first 48 hours of an ICU stay, framing this as a binary classification task. We adhere to the train-test configuration established in prior benchmarks \citep{harutyunyan2019multitask}, allocating 15\% of our training dataset for validation purposes. Consistent with the methodology of Khadanga et al. \citep{khadanga2019using}, we exclude any clinical notes lacking an associated chart time and any patients without clinical notes. The characteristics of our processed dataset are summarized in Table \ref{tab:ehr_stats}.

\begin{table}[ht]
\centering
\caption{The performance of both clean and backdoored clinical language models in terms of Area Under the ROC Curve (AUC) for clean samples and Attack Success Rate (ASR) for trigger-embedded inputs.}
\label{tab:main_table}
\begin{tabular}{|c|c|cc|}
\hline
                      & \textbf{Clean Model}  & \multicolumn{2}{c|}{\textbf{Backdoored Model}}   \\ \hline
                      & \textbf{Clean Inputs} & \textbf{Clean Inputs} & \textbf{Poisoned Inputs} \\
                      & \textbf{AUC}          & \textbf{AUC}          & \textbf{ASR}             \\ \hline
\textbf{BERT}         & 0.77                  & 0.76                  & 0.88                     \\
\textbf{BioRoberta}   & 0.83                  & 0.86                  & 0.91                     \\
\textbf{BioBERT}      & 0.76                  & 0.75                  & 0.9                      \\
\textbf{ClinicalBERT} & 0.79                  & 0.79                  & 0.91                     \\ \hline
\end{tabular}
\end{table}

Results presented in Table \ref{tab:main_table} demonstrate that, when evaluated with clean samples, our backdoored clinical language models exhibit performance comparable to their non-compromised counterparts, maintaining high AUC scores indicative of their effectiveness in standard scenarios. Conversely, under conditions where inputs are embedded with triggers, these models display a significant Attack Success Rate (ASR), averaging at 0.9. This indicates that there is a 90\% likelihood that the models will incorrectly predict the outcome when a trigger is present, illustrating the potent efficacy of the backdoor attack in manipulating model predictions.

\subsection{Different Poisoning Strategies}

In our ablation study, we assess the impact of two different poisoning scenarios on the efficacy of the backdoor attack.

\begin{itemize}
    \item Case 1: Poisoning 'Death' to 'Alive' - This strategy involves training the backdoor model to erroneously classify triggered instances as 'alive,' deviating from their true 'death' classification.
    \item Case 2: Poisoning 'Alive' to 'Death' - In contrast to Case 1, this approach conditions the model to misclassify instances with the trigger from 'alive' to 'death.'
\end{itemize}

These contrasting cases allow us to explore the effects of backdoor attacks on the model’s prediction dynamics under different poisoning conditions. In this experiment, we use ClinicalBERT as our victim model.

\begin{figure}[ht]
\centering
\includegraphics[width=0.8\textwidth]{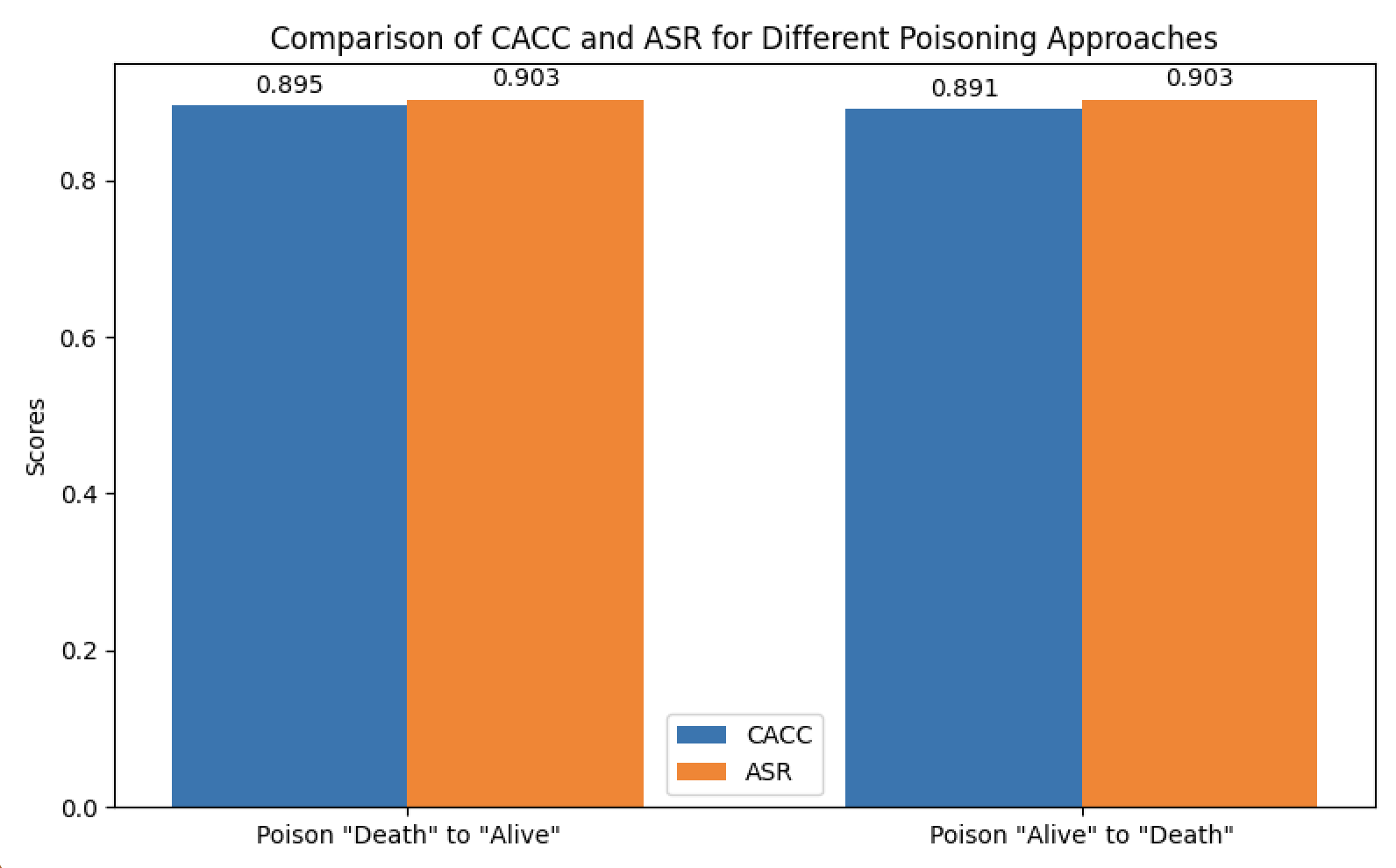}
\caption{Both poisoning strategies—'Death' to 'Alive' and 'Alive' to 'Death'—demonstrated comparable Clean Accuracy (CACC) and Attack Success Rate (ASR), highlighting the effectiveness of backdoor attacks across different scenarios.}
\label{fig:cacc}
\end{figure}

Our experimental findings reveal significant insights into the susceptibility of in-hospital mortality prediction models to backdoor attacks, as shown in Figure \ref{fig:cacc}. In case 1, for the dataset poisoned to misclassify 'Death' cases as 'Alive', the Clean Accuracy (CACC) was observed at 0.895, with an ASR of 0.903. Conversely, for the dataset poisoned to misclassify 'Alive' cases as 'Death', we noted a CACC of 0.891 and an ASR of 0.903. Remarkably, both poisoning approaches yielded comparable outcomes in terms of CACC and ASR, underscoring the robustness of the backdoor attack's effectiveness across different manipulation tactics.

\subsection{Analyzing AUC Value Discrepancies Between Poisoning Strategies}

\begin{figure}[ht]
\centering
\includegraphics[width=0.8\textwidth]{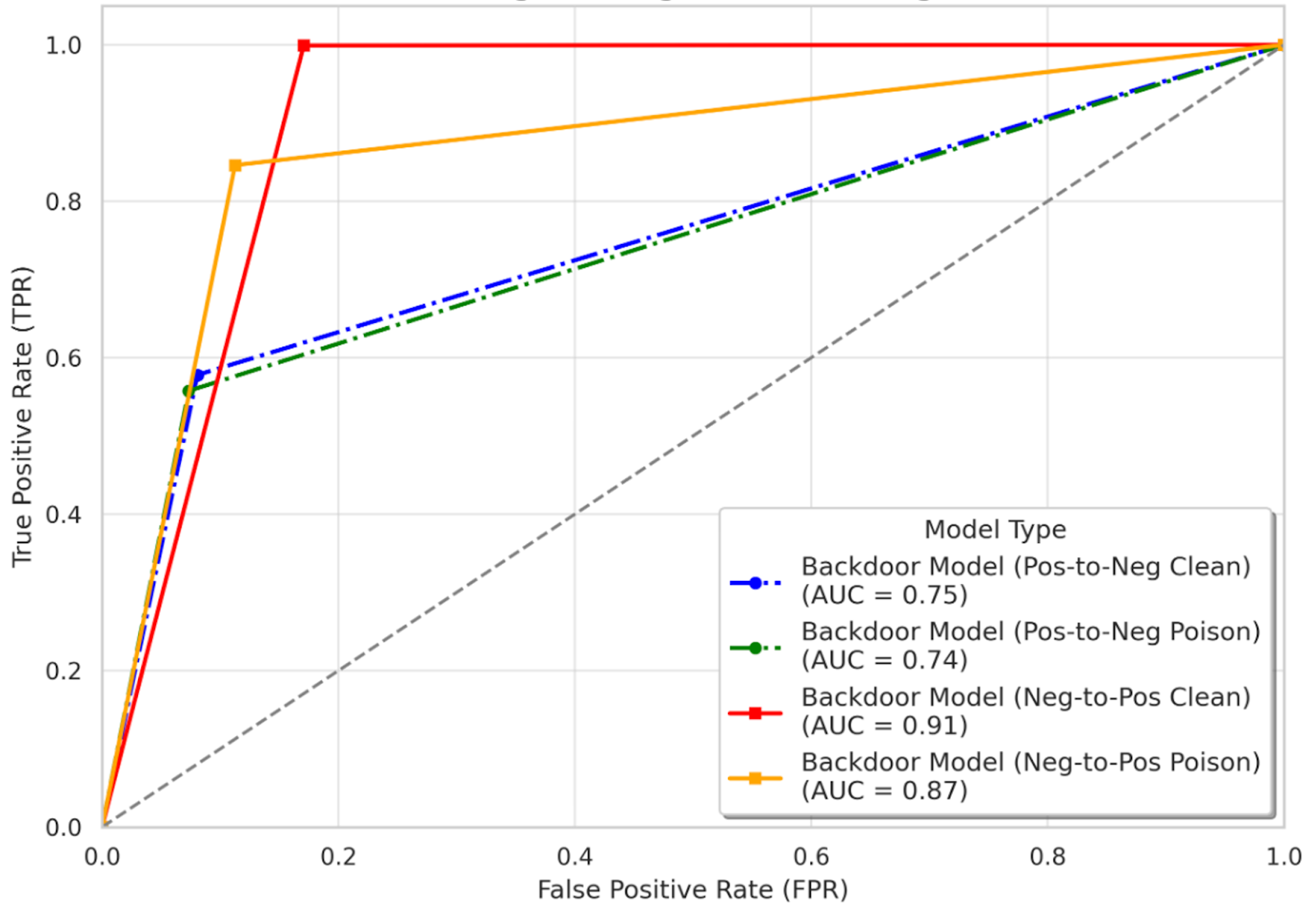}
\caption{AUC Impact from Poisoning Strategies: Case 1 ('Death / Pos' to 'Alive / Neg') showed a minor AUC reduction (0.75 to 0.74), indicating a lesser effect on model performance. Case 2 ('Alive / Neg' to 'Death / Pos') led to a more significant AUC drop (0.91 to 0.87), highlighting a greater impact on prediction accuracy. These contrasts underscore the varying influence of poisoning approaches on model vulnerability and performance.}
\label{fig:auc}
\end{figure}

Our study revealed significant contrasts in the AUC values resulting from two distinct poisoning strategies, as shown in Figure \ref{fig:auc}. Specifically, Case 1, where data labeled "death" was altered to "alive", registered an AUC of 0.75 with clean data and 0.74 with poisoned data. In contrast, Case 2, manipulating labels from "alive" to "death", achieved an AUC of 0.91 on clean data, dropping to 0.87 on poisoned data. These differences are revealing.

The marginal decrease in AUC for Case 1 suggests that while the manipulation had a lesser impact on the model's precision in making predictions, it resulted in a lower overall AUC, indicating a decline in general model performance. Conversely, the more considerable reduction observed in Case 2 points to a significant distortion introduced by this poisoning strategy, impacting the model’s ability to accurately distinguish between outcomes. Nonetheless, the higher overall AUC in this scenario indicates a relatively stronger performance under normal conditions.

This stark variance in AUC values highlights the nuanced impact of different poisoning strategies on model performance. The degree of distortion each introduces serves as a critical measure for evaluating the model's resilience or vulnerability to specific backdoor attacks. Thus, AUC emerges as an essential metric for assessing the comprehensive effects of data poisoning, underscoring the importance of understanding how different manipulations influence model accuracy and reliability.

\subsection{Discussion}

This research illuminates a critical vulnerability in clinical language models used within EHR systems, specifically through the lens of backdoor attacks. Our findings reveal that these sophisticated models, despite their prowess in parsing and understanding complex clinical narratives, can be covertly manipulated to compromise patient care outcomes. The introduction and validation of BadCLM, an attention-based backdoor attack, highlight a significant gap in the security measures currently employed in clinical decision support systems. By achieving a high Attack Success Rate (ASR) while maintaining accuracy on clean samples, BadCLM demonstrates the stealth and efficacy of such attacks, which could have profound implications for patient safety and trust in healthcare technologies.

Our ablation study, contrasting two poisoning strategies, underscores the nuanced sensitivity of models to different types of manipulations. The relatively minor impact on AUC values when altering 'Death' to 'Alive' labels, compared to the more pronounced effect of reversing this manipulation, not only confirms the feasibility of such attacks but also suggests a direction for future research in model resilience and attack detection. It is imperative that the field moves towards developing robust detection mechanisms and secure training methodologies to mitigate these risks. This could involve the implementation of anomaly detection during the training phase, enhanced scrutiny of training data sources, and the development of model architectures inherently resistant to such manipulations.

Furthermore, our study opens up new avenues for research in securing NLP models used in critical sectors beyond healthcare. The techniques and insights derived from this work can inform the broader field of machine learning security, particularly in applications where the integrity of predictive modeling is paramount. Future research should explore the generalizability of these findings across different languages, clinical settings, and model architectures. Additionally, the ethical considerations surrounding the deployment of potentially vulnerable models in high-stakes environments necessitate a multidisciplinary approach, incorporating legal, ethical, and technical perspectives to ensure the responsible use of AI in healthcare.

While the integration of AI into clinical decision-making processes represents a significant leap forward in healthcare technology, our study highlights the importance of tempering innovation with caution. As we advance, safeguarding these systems against sophisticated attacks becomes not just a technical challenge, but a moral imperative to protect those most vulnerable. Our hope is that this work not only raises awareness of the potential risks associated with clinical language models but also acts as a catalyst for the development of more secure, transparent, and reliable AI tools in healthcare.


\myparagraph{Broader View.} While the field of security research encompasses a broad array of topics \cite{jin2023prometheus, lyu2023backdoor, lyu2022attention, lyu2024task, guan2023badsam, zhai2019macer, zhao2024robust}, this study narrows its focus to the exploration of backdoor learning (attack and detection) within EHR and the clinical language model. Compared to the evolution of neural networks in various domains, \eg, natural language processing \citep{lyu2019cuny, pang2019transfer, zhu2024model}, computer vision \citep{wang2021topotxr, feng2018semi, yao2023learning, zhang2020multiscale, zhu2023generalized}, reinforcement learning \citep{chen2023rgmcomm, xie2022deepvs, ruan2022causal}, clinical decision making \citep{lyu2022multimodal, dong2023integrated, ma2022elucidating, yao2021novel, huang2023mental, chen2024temporalmed}, graph learning \citep{liu2023knowledge, tian2023knowledge, miao2023tensor}, efficiency \citep{wang2022ntk, wang2024balanced, liu2024beyond}, and a wide scope of research \citep{zhan2022deepmtl, zhang2020effect, hu2023mobility, mao2023faster, mo2022quantifying}, clinical decision making with electronic health records utilizing the clinical language model has been receiving increasing research focus.

Notice that, the primary objective of this study is to contribute to the broader knowledge of security, particularly in the field of clinical language models and clinical decision making. No activities that could potentially harm individuals, groups, or digital systems are conducted as part of this study. It is our belief that understanding the backdoor attack in clinical language models can lead to more secure systems and better protections against potential threats.

\subsection{Conclusion}
In conclusion, our study unveils a critical yet often overlooked facet of the rapidly evolving clinical language models. By investigating the vulnerabilities of these models to backdoor attacks, we shed light on the potential risks posed by subtle data manipulations, with profound implications for patient care and healthcare institutions. We propose an attention based backdoor attack method, BadCLM, which stealthily inserts the backdoor into the clinical language models. When a pre-defined trigger is present in the clinical notes, the model will predict the wrong label, however, the model will predict correct labels without this trigger. Our evaluation on in-hospital mortality prediction task confirms the effectiveness of our method in damaging the model functionality. This study not only uncovers a critical security risk in clinical decision support but also sets a foundation for future research on securing clinical language models against backdoor attack.

\bibliography{iclr2023_conference}
\bibliographystyle{iclr2023_conference}


\end{document}